\documentclass[lettersize,journal]{IEEEtran}
\usepackage{amsmath,amsfonts}
\usepackage{amsthm}
\usepackage{array}
\usepackage[caption=false,font=normalsize,labelfont=sf,textfont=sf]{subfig}
\usepackage{textcomp}
\usepackage{stfloats}
\usepackage{url}
\usepackage{verbatim}
\usepackage{graphicx}
\usepackage{cite}
\usepackage{xcolor}
\usepackage{hyperref}
\usepackage{url}
\usepackage{here}
\usepackage{algorithm}
\usepackage{amssymb}

\usepackage[noend]{algpseudocode}
\algrenewcommand\algorithmicrequire{\textbf{Initiate}}
\usepackage{tabularx}
\usepackage[bottom]{footmisc}
\usepackage{comment}
\usepackage{nomencl}

\usepackage{adjustbox}
\usepackage{subcaption}
\usepackage{overpic}
\usepackage{tabularx}
\usepackage{array}
\newcolumntype{Y}{>{\raggedright\arraybackslash}X}
\usepackage{graphicx}
\usepackage{overpic}
\usepackage{xcolor}
\usepackage{subcaption}
\usepackage{booktabs}
\usepackage{multirow}

\title{Perceptual Comparisons of Viscoelastic Renderings: Fractional- vs Integer-Order Models}

\begin{document}

\author{Gorkem Gemalmaz \qquad \qquad \qquad \qquad Harun Tolasa \qquad \qquad \qquad \qquad Volkan Patoglu\thanks{G. Gemalmaz, H. Tolasa and V. Patoglu are with the Faculty of Engineering and Natural Sciences at Sabanci University,  Istanbul, Turkiye.\\
    {\tt\scriptsize \{gorkem.gemalmaz,harun.tolasa,vpatoglu\}@sabanciuniv.edu}}
}

\maketitle

\begin{abstract}
Realistic haptic rendering of viscoelastic materials requires reproducing time-dependent behaviors such as creep and stress relaxation, which arise from the material's inherent memory effects. Fractional-order viscoelastic models have been shown to capture these dynamics accurately and compactly, but it remains unclear whether their increased fidelity translates into perceptual benefits during haptic interaction. In this work, we investigate whether users perceive fractional-order models as more realistic than conventional integer-order models while rendering viscoelastic materials. We compare the fractional-order standard linear solid~(FO-SLS) model with two integer-order alternatives: the 3-parameter SLS model and the 5-parameter generalized Maxwell model. We consider two complementary approaches to optimal parameter selection to evaluate rendering realism from both objective and perceptual perspectives: In the first approach, we determine the model parameters via least-squares based system identification of physical viscoelastic reference materials to establish evaluation baselines.  In the second approach, we determine population-level perceptually optimal parameters using a qualitative-feedback-based human-in-the-loop optimization and aggregation method. This method establishes model parameters that maximize perceived realism of viscoelastic behaviors through qualitative evaluations. Human-subject experiments indicate that the FO-SLS model is perceived as significantly more realistic than both integer-order alternatives under both parameter selection approaches. Overall, this study provides experimental evidence that fractional-order viscoelastic models are not only theoretically well-motivated, but also perceptually relevant for realistic haptic rendering of creep- and relaxation-dominated behaviors of viscoelastic materials.
\end{abstract}

\begin{IEEEkeywords}
Physical human-robot interaction, haptic rendering of viscoelasticity, fractional-order standard linear solid model, human-in-the-loop optimization, and perceived realism.
\end{IEEEkeywords}

\section{Introduction}

Viscoelastic materials, including biological tissues and polymers, show inherently time-dependent mechanical behavior. In particular, these materials exhibit phenomena such as creep and stress relaxation, which arise from the dependence of the response on the deformation and loading history. Accurately capturing these behaviors is important for applications involving physical interaction with soft materials, particularly in haptic medical and surgical training simulators~\cite{salisbury1995haptic}. For instance, palpation can detect changes in tissue viscoelasticity associated with abnormalities such as tumors, and haptic medical simulators may assist in training for diagnosing such tumors~\cite{breast_cancer,breast_tissue, prostate}. Therefore, realistic rendering of viscoelastic tissue behavior is essential for applications involving physical interaction with soft biological materials. 

Selecting an appropriate model for viscoelastic haptic rendering requires considering both how accurately the model captures the physical material response and how realistic the resulting interaction is perceived by users. Classical compliance models based on integer-order~(IO) spring-damper networks, including the standard linear solid~(SLS) model and the generalized Maxwell~(GM) model, approximate these behaviors, but fail to represent long-term memory effects and frequency-dependent dynamics accurately~\cite{breast_tissue}. Fractional-order~(FO) viscoelastic models provide an appealing alternative, as non-integer derivatives naturally encode memory-dependent dynamics and can represent complex viscoelastic responses with relatively few model parameters~\cite{FO_few_param, BagleyTorvik1983,mainardi2010}.

Despite these modeling advantages, it remains unclear whether the improved descriptive performance of FO models necessarily translates into \emph{perceptually more realistic} haptic rendering. Model accuracy is commonly evaluated through physical characterization, by assessing how well a model fits measured response in a least-squares sense. On the other hand, during haptic rendering, an accurate physical description of the reference does not necessarily correspond to the most realistic percept during physical interactions~\cite{Swindells2009,Zhang2022}.

This distinction is particularly relevant for viscoelastic rendering, where the perceived response depends on the combined perception of stiffness, damping, and the interaction history provided by the haptic interface.  Consequently, comparing viscoelastic rendering models only through system-identification error or model complexity is insufficient to determine which model provides the most realistic haptic experience. A fair comparison among viscoelastic rendering models requires not only determining optimal parameters for each model to capture the same physical reference material, but also evaluating whether users perceive the resulting renderings as realistic.

Determination of optimal parameters can be achieved through system identification or human-in-the-loop~(HiL) optimization of perceived realism~\cite{Catkin2023, Tolasa_Pareto, Tolasa2026}. While tuning the model parameters through HiL optimization, different model structures introduce different trade-offs. A lower-dimensional IO model is relatively easier to tune and interpret, whereas a higher-dimensional IO model can represent richer dynamics at the cost of a larger parameter space. An FO model can capture memory-dependent behavior more compactly with fewer parameters, but the coupling among its parameters makes their individual perceptual effects harder to predict.

In this study, we investigate whether the modeling differences and trade-offs between FO and IO viscoelastic models are also reflected in their perceived realism during haptic rendering. 
In particular, we evaluate whether the compact representation of memory-dependent behavior provided by the FO-SLS model results in higher perceived realism than its IO counterparts, and whether increasing the number of parameters in an IO model provides sufficient improvement in perceived realism to compensate for the added complexity.

Accordingly, we compare the rendering performance of the 4-parameter FO-SLS model with two integer-order alternatives of different model complexities: the 3-parameter IO-SLS model and the 5-parameter IO-GM model. We perform such comparisons under two complementary optimal parameter-selection approaches. First, we compare model parameters identified through physical characterization of the reference viscoelastic materials. Second, we compare population-level perceptually optimal parameters determined from participants' qualitative feedback through HiL optimization experiments conducted according to~\cite{Tolasa2026}. This way, we evaluate the effects of model structure and complexity on perceived realism under both physically and perceptually optimized parameter sets.

The main contribution of this paper is a human-subject comparison of FO and IO viscoelastic rendering models under various optimal parameter selection settings. By comparing the FO model with IO models of increasing complexity, using qualitative participant evaluations, we provide experimental evidence indicating that rendering viscoelasticity with FO-SLS models results in perceptually more realistic interactions than conventional IO alternatives. Overall, this study shows that FO models are not only theoretically more suitable for describing viscoelastic materials but also that users perceive these modeling advantages while rendering viscoelasticity.

\vspace{1mm}
\section{Related Work}

\subsection{Viscoelastic Models for Haptic Rendering}

Creep and stress relaxation arise from internal memory~\cite{time_dependant_tissue}: creep captures gradual deformation under constant stress, while stress relaxation models force decay under constant strain. These effects are crucial for capturing the response of soft biological tissues and polymeric materials and cannot be adequately represented by purely elastic models~\cite{IO_fails_memory}. Common IO viscoelastic representations include the Kelvin--Voigt, Maxwell, SLS, and GM models. These models differ in their ability to capture creep and relaxation behavior.

\begin{figure}[t!]
  \includegraphics[width=.5\textwidth]{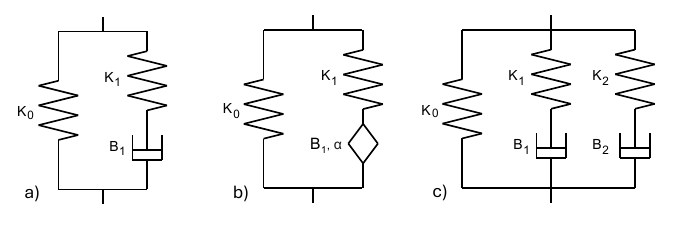}  
  \vspace{-5mm}
  \caption{
    Viscoelastic models used in the experiments:
    (a)~standard linear solid (SLS) model with 3 parameters,
    (b)~fractional-order~(FO) SLS model with 4 parameters, and
    (c)~generalized Maxwell (GM) model with 5 parameters.}
\label{fig:visco_models}\vspace{-1mm}
\end{figure}

Classical IO viscoelastic models are easier to interpret and computationally convenient since they are built from networks of ideal springs and dampers. The Kelvin-Voigt model can capture bounded creep, but is inadequate to represent stress relaxation~\cite{kelvin_orj,voigt_orj, kelvin}, while the Maxwell model captures stress relaxation but results in unbounded creep~\cite{maxwell_orj,kelvin-maxwell}. The SLS model~(Fig.~\ref{fig:visco_models}(a)) improves on these limitations by more accurately representing both creep and stress relaxation; however, its single relaxation time constant limits its fidelity when modeling materials with broader relaxation spectra~\cite{zener_orj,frac_sls}.  GM models~(Fig.~\ref{fig:visco_models}(c)) further increase expressiveness at a cost of model complexity by adding several Maxwell branches with different relaxation time constants~\cite{wiechert_orj,frac-intg_model}; however, a finite number of such branches cannot fully account for the history dependence underlying viscoelasticity~\cite{wiechert_orj,frac-intg_model}.

Fractional-order~(FO) viscoelastic models~(Fig.~\ref{fig:visco_models}(b)) offer a compact alternative by replacing classical damping with a fractional-order one. Since fractional derivatives depend on the full history of the signal, FO models can naturally capture memory-dependent behavior, such as intermediate viscoelastic responses between solid-like and fluid-like behaviors in materials exhibiting creep and relaxation characteristics~\cite{frac_sls,cagatay_liver,frac_better}.  For instance, the FO-SLS model can model broad relaxation behavior that would otherwise require infinitely many IO branches in the GM model; that is, the GM model converges to an FO-SLS model in the limit, as the number of Maxwell branches approaches infinity~\cite{kelvin,gmm-sls}. This property makes the FO-SLS model attractive for haptic rendering, where realistic time-dependent behavior must be captured with a manageable number of model parameters~\cite{cagatay_liver,tokatli2015,tokatli_patoglu_2018}.

On the other hand, theoretical modeling advantages do not necessarily translate into higher perceived realism during physical interactions with a haptic interface. A low-parameter IO model, as in Fig.~\ref{fig:visco_models}(a), is easier to tune; a higher-parameter IO model, as in Fig.~\ref{fig:visco_models}(c), may approximate richer dynamics with additional parameters, while an FO model, as in Fig.~\ref{fig:visco_models}(b), can capture memory effects compactly but with less intuitive parameters that display frequency-dependent couplings. 

\vspace{-2mm}
\subsection{Perceptual Evaluation of Viscoelastic Haptic Rendering}

Perceived realism is commonly evaluated through human judgments of similarity, preference, or naturalness with respect to interactions with a physical reference object~\cite{Culbertson2017, Turchet2013, Friesen2024, Catkin2023, Tolasa_Pareto, Tolasa2026}. Studies in the literature indicate that physical model accuracy alone is insufficient to evaluate the realism of haptic rendering, and that when the target outcome is the \emph{perceived realism}, human evaluation constitutes the gold standard~\cite{Rosenberg1993}. 

Qualitative user evaluations have therefore become an important complement to quantitative physical characterizations. Data-driven rendering studies have shown that while measured interactions can capture complex responses, perceptual validations are still required to determine whether the rendered interaction feels similar to the physical material~\cite{Hover2010,Yim2016}. 

This is particularly relevant for rendering viscoelasticity, since the perception of the material depends on multiple exploratory cues that are time-dependent, including stiffness, damping, creep, relaxation, and other kinesthetic force-motion relationships~\cite{Srinivasan1995, BergmannTiest2009, Caldiran2019}. Psychophysics studies have established systematic means of evaluation when the quantity of interest is perceived similarity or realism~\cite{Jones2013, Okamoto2013, Catkin2023, Tolasa2024, Tolasa2026}. 

Earlier works on the perceptual evaluation of haptic rendering of viscoelasticity
have focused on physical modeling with perceptual experiments, and relied on user-centered evaluation of realism to verify the results~\cite{Cha2022, Tolasa2026, Caldiran2018, Caldiran2019}. However, none of these related works addressed whether different viscoelastic model families produce different perceived realism for the same material. In this study, we address this gap by directly comparing the FO-SLS models with low- and high-parameter IO models, using both physically characterized and population-level perceptually optimized parameters for each model. Accordingly, we test whether users can perceive the modeling advantages of FO models while rendering viscoelastic behavior.

\section{Preliminaries}
\label{sec:Preliminaries}
\subsection{Fractional-Order Model of Viscoelasticity}

In this study, we use the fractional derivative defined according to the Grünwald--Letnikov~(GL) definition~\cite{grunwald}, which is well suited for discrete-time implementation of FO models. According to the GL definition, given a function~$f(x)$, its derivative of order $\alpha$ is given by
\begin{equation}
D^\alpha f(x) = \,
\lim_{h \to 0} \, \frac{1}{h^\alpha} \,
\sum_{k=0}^{\infty} \, (-1)^k \, \binom{\alpha}{k}\,f(x-kh)
\nonumber
\end{equation}

\noindent where $\alpha \in (0,1)$, $h$ is the step size, with

\begin{equation}
\binom{\alpha}{k}
=
\frac{\Gamma(\alpha+1)}
{\Gamma(k+1) \, \Gamma(\alpha-k+1)}.
\nonumber
\end{equation}

Unlike an IO derivative, this operator depends on all previous signal values through a weighted sum of the history. Therefore, a fractional-order element provides a direct mechanism for capturing memory effects.

Fractional-order differentiation is a linear operator. Assuming zero initial conditions, the fractional derivative corresponds to multiplication by $s^\alpha$ in the Laplace domain. Accordingly, the force-position impedance transfer function of the FO-SLS model can be expressed as

\begin{equation}
H_{FO-SLS}(s)=K_0+\frac{K_1 B_1 s^\alpha}{K_1+B_1 s^\alpha},
\nonumber
\end{equation}

\noindent where $K_0$ and $K_1$ are the stiffness coefficients, $B_1$ is the fractional damping coefficient, and $\alpha$ determines the relative elastic-viscous behavior of the fractional element. In this study, we compare this model with the 3-parameter IO-SLS and 5-parameter IO-GM models to determine whether FO representation leads to higher perceived realism. 

\bigskip
\subsubsection*{Short-Memory Discretization}

As the GL definition shows, an exact implementation of FO dynamics requires infinite memory, which is infeasible for real-time control. Therefore, we use the short-memory approximation to discretize the model and provide a practical discrete-time realization, while preserving the memory-dependent behaviors~\cite{short_memory}. In short-memory discretization, we truncate the history term of the GL derivative to the most recent \(N\) samples, since more recent samples dominate the weighted sum.

In particular, given a discrete-time signal \(x(t)\), the fractional derivative of order \(\alpha\) is discretized in the \(z\)-domain as  \vspace{-2mm}

\begin{equation}
\mathcal{Z}\{D^\alpha x(t)\}
\approx
\frac{1}{T^\alpha}
\sum_{i=0}^{N}
c_i z^{-i},
\nonumber
\end{equation}

\noindent where \(T\) is the sampling period, \(N\) is the memory window length, and the GL coefficients are computed recursively as

\vspace{-2mm}
\begin{equation}
c_0 = 1, \qquad
c_i =
\frac{i-\alpha-1}{i}c_{i-1},
\quad i=1,\ldots,N.
\nonumber
\end{equation}

\noindent For \(\alpha=1\), this formulation reduces to the standard backward-difference discretization.

After applying the short-memory discretization, the discrete-time force-position impedance of the FO-SLS model becomes

\begin{equation*}
\label{eq:SLS}
H_{FO-SLS}(z)=K_0+\frac{K_1 B_1 T^{-\alpha} \sum_{i=0}^{N} c_i z^{-i}}{K_1+B_1 T^{-\alpha} \sum_{i=0}^{N} c_i z^{-i}}.
\end{equation*}

Here, \(N\) controls the retained memory length and therefore affects how accurately the rendered model reproduces time-dependent viscoelastic behavior. In this study, we set $N$ to 101 to provide passive, realistic viscoelasticity renderings while ensuring computational feasibility, as established in~\cite{Tolasa2026,Gemalmaz2026}.

\subsection{Integer-Order Models of Viscoelasticity}

To evaluate whether the FO model provides a perceptual advantage during haptic rendering, we compare the FO-SLS model with two IO alternatives: the 3-parameter IO-SLS model and the 5-parameter IO-GM model, which serve as low- and high-parameter IO baselines, respectively. The IO-SLS model has the same spring-damper structure as the FO-SLS model, but replaces the fractional-order element with a classical damper. The IO-GM model increases the number of IO Maxwell branches to two to capture a broader range of relaxation behavior.

\bigskip
\paragraph{IO-SLS Model}

The 3-parameter IO-SLS model, with force-position impedance transfer function

\begin{equation*}
H_{SLS}(s)=K_0+\frac{K_1 B_1 s}{K_1 + B_1 s}
\end{equation*}

\noindent serves as the low-parameter IO model. In this model,  \(K_0\) and \(K_1\) denote the stiffness coefficients, while \(B_1\) is the damping coefficient. 

For modeling consistency, backward-difference discretization is used for IO models, since GL derivative reduces to backward-difference when \(\alpha=1\). Using the sampling period \(T\), the corresponding discrete-time impedance transfer function reads as

\begin{equation*}
H_{SLS}(z)=\frac{(B_1K_1 + B_1K_0 + K_1K_0T)z-(B_1K_1 + B_1K_0)}{(B_1 + K_1T)z - B_1}.
\end{equation*}

This model provides an interpretable, low-dimensional baseline for assessing whether FO models improve perceived realism of rendering.

\medskip
\paragraph{IO-GM Model}

As an alternative, more expressive IO model, we also consider a 5-parameter GM model with two Maxwell branches in parallel to a static stiffness. The force-position impedance transfer function of this model is 

\begin{equation*}
H_{GM}(s)=K_0+\frac{K_1 B_1 \, s}{K_1 + B_1 \, s}+\frac{K_2 B_2 \, s}{K_2 + B_2 \, s}
\end{equation*}

\noindent where \(K_0\) is the parallel stiffness, and \((K_1,B_1)\) and \((K_2,B_2)\) denote the stiffness and damping elements of the two Maxwell branches. The additional branch allows the model to capture multiple relaxation time scales, making it more expressive than the SLS~model baseline. The discrete-time impedance transfer function of this model is given as

\begin{equation*}
H_{GM}(z)=\frac{A\,z^2 + B\,z + C}{D}
\end{equation*}

\vspace{-3mm}
\noindent with \vspace{-3mm}

\begin{equation}
\resizebox{\columnwidth}{!}{$
\begin{aligned}
A &= B_1 B_2 (K_0 + K_1 + K_2)
+ T\Big[B_1 K_2 (K_0 + K_1)
+ B_2 K_1 (K_0 + K_2)\Big]
+ K_0 K_1 K_2 T^2, \\[6pt]
B &= -2 B_1 B_2 (K_0 + K_1 + K_2)
- T\Big[B_1 K_2 (K_0 + K_1)
+ B_2 K_1 (K_0 + K_2)\Big], \\[6pt]
C &= B_1 B_2 (K_0 + K_1 + K_2), \\[6pt]
D &= (B_1 + K_1 T)(B_2 + K_2 T)\,z^2
- \Big[2B_1B_2 + T(B_1K_2 + B_2K_1)\Big]\,z
+ B_1B_2 .
\end{aligned}
$}
\nonumber
\end{equation}

Comparing the IO-GM model with the FO-SLS model helps assess whether adding extra parameters to IO models can match the perceptual realism achieved by inherently more compact FO representations.

\section{Determination of Model Parameters for Viscoelasticity Rendering}
\label{sec:param_determination}

\begin{figure*}[t!]
    \centering
    \includegraphics[width=1\textwidth]{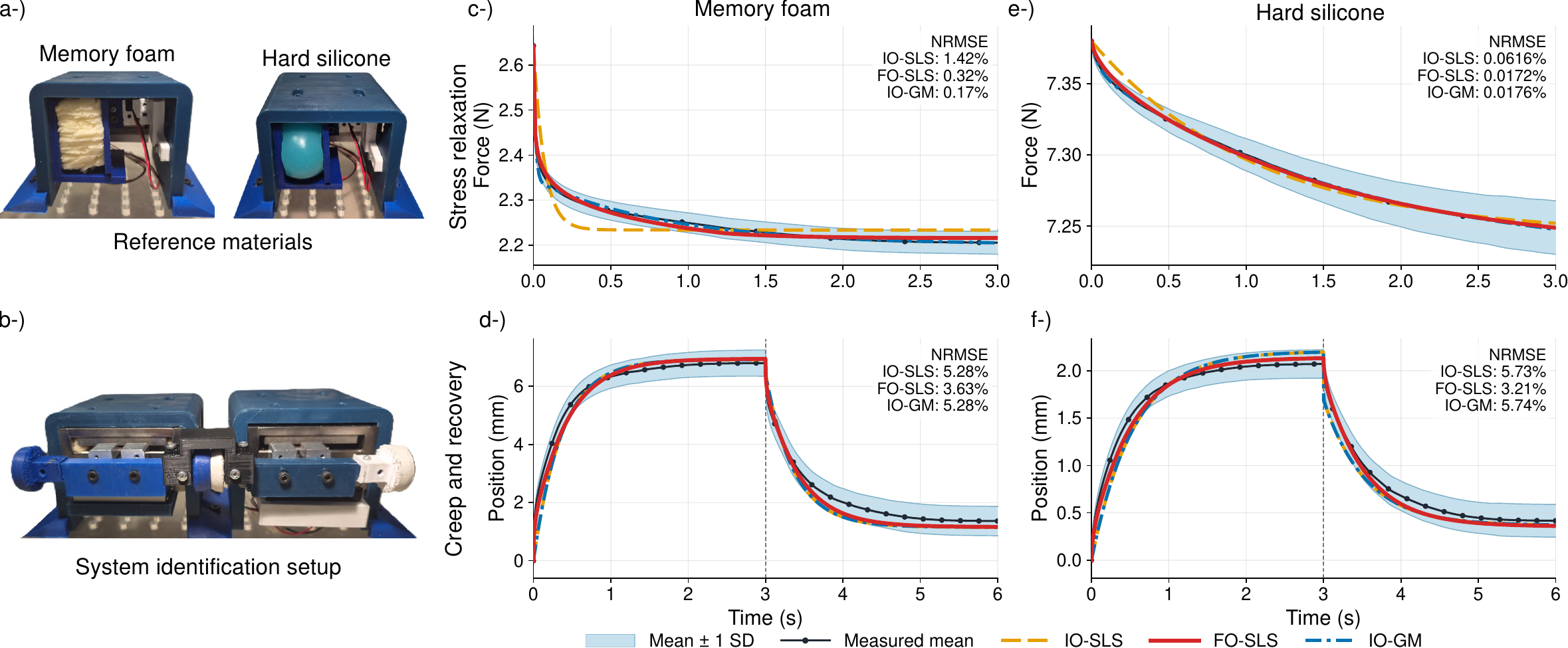}
    \vspace{-5.5mm}
    \caption{(a)~Reference materials used in the experiment. (b)~System identification setup. (c)--(f)~Characterization experiment data and best fits of IO-SLS, IO-GM, and FO-SLS rendering models. The top row shows the results for stress relaxation, while the bottom row shows the results for creep-with-recovery experiments. Experiments in (c)~and~(d) were conducted with a memory foam reference, while (e)~and~(f) were conducted with a hard silicone material.} 
    \label{Fig:SysID_combined}
\end{figure*}

A fair comparison among multiple viscoelasticity models requires determining \emph{optimal} parameters for each model. One means to determine optimal parameters is to rely on physically grounded parameters from system identification experiments. Determining parameters through system identification experiments can quantitatively ensure the best match to the measured system response. On the other hand, strong evidence in the literature shows that system-identified parameters do not necessarily yield the best perceived realism in haptic rendering. An alternative way to determine optimal parameters is to perform HiL optimization experiments based on qualitative user feedback from a population of participants and aggregate these results to estimate population-level \emph{perceptually optimal} model parameters that yield the highest perceived realism during psychophysical evaluations~\cite{Catkin2023, Tolasa2024, Tolasa2026}. 

In this study, we utilize both (i)~physically grounded optimal parameters from least-squares-based system identification and (ii)~population-level perceptually optimal parameters from the aggregation of the HiL optimization results.

For all viscoelasticity models, the feasible parameter sets were constrained to satisfy the corresponding passivity conditions of the sampled-data control system to ensure coupled stability of interactions during rendering~\cite{Gemalmaz2026}. In addition, we matched the effective stiffness of each virtual environment to that of the physical reference in the low-frequency region, so the comparison emphasized differences in time-dependent viscoelastic behavior, such as creep and stress relaxation, rather than more trivial differences in overall stiffness. 

To evaluate the viscoelastic rendering models, we used two physical references with distinct material properties to verify whether the findings extend across different viscoelastic interactions, as shown in Fig.~\ref{Fig:SysID_combined}~(a). In particular, memory foam exhibits evident history-dependent deformation and recovery, while hard silicone produces a more stiffness-dominated compression and release response. These two reference materials capture two representative extremes of a broad range of viscoelastic behaviors.

\begin{table*}[t!]
\centering \vspace{1mm}
\caption{Parameters determined via system identification~(SysID) and population-level perceptually optimized parameters from HiL experiments~(HiL). SysID NRMSEs are computed from the stress-relaxation fits, while normalized errors denote parameter-wise distances between the corresponding SysID and HiL parameter values relative to the search-space ranges. The error for $K_0$ is not reported since it is determined via effective stiffness matching.}
\label{tab:hil_sysid_combined}
\renewcommand{\arraystretch}{1.2}
\footnotesize

\begin{tabular*}{\textwidth}
{@{\extracolsep{\fill}} l cccccc cccccc @{}}

& \multicolumn{6}{c}{\textbf{Memory Foam}}
& \multicolumn{6}{c}{\textbf{Hard Silicone}} \\
\hline\hline

\multicolumn{13}{c}{\textbf{FO-SLS (4-parameter model)}} \\
\hline

\textbf{Set}
& \(K_0\) & \(K_1\) & \(B_1\) & \(\alpha\) & & NRMSE
& \(K_0\) & \(K_1\) & \(B_1\) & \(\alpha\) & & NRMSE \\
\hline

HiL
& \(-3.18\) & \(5.58\) & \(7.88\) & \(0.18\) & & --
& \(-2.19\) & \(4.97\) & \(7.59\) & \(0.53\) & & -- \\

SysID
& \(-2.89\) & \(5.70\) & \(5.89\) & \(0.20\) & & \(0.32\%\)
& \(-1.88\) & \(4.72\) & \(6.35\) & \(0.61\) & & \(0.0172\%\) \\

Norm.\ error
& -- & \(0.40\%\) & \(6.20\%\) & \(2.80\%\) & & --
& -- & \(0.82\%\) & \(3.88\%\) & \(7.89\%\) & & -- \\

\hline\hline

\multicolumn{13}{c}{\textbf{IO-SLS (3-parameter model)}} \\
\hline

\textbf{Set}
& \(K_0\) & \(K_1\) & \(B_1\) & & & NRMSE
& \(K_0\) & \(K_1\) & \(B_1\) & & & NRMSE \\
\hline

HiL
& \(0.44\) & \(1.00\) & \(0.03\) & & & --
& \(1.45\) & \(1.99\) & \(0.067\) & & & -- \\

SysID
& \(0.44\) & \(1.06\) & \(0.01\) & & & \(1.42\%\)
& \(1.45\) & \(0.88\) & \(0.078\) & & & \(0.0616\%\) \\

Norm.\ error
& -- & \(0.18\%\) & \(0.06\%\) & & & --
& -- & \(3.58\%\) & \(0.03\%\) & & & -- \\

\hline\hline

\multicolumn{13}{c}{\textbf{IO-GM (5-parameter model)}} \\
\hline

\textbf{Set}
& \(K_0\) & \(K_1\) & \(B_1\) & \(K_2\) & \(B_2\) & NRMSE
& \(K_0\) & \(K_1\) & \(B_1\) & \(K_2\) & \(B_2\) & NRMSE \\
\hline

HiL
& \(0.44\) & \(1.05\) & \(0.03\) & \(1.43\) & \(0.0010\) & --
& \(1.45\) & \(1.00\) & \(0.0062\) & \(1.78\) & \(0.0575\) & -- \\

SysID
& \(0.44\) & \(0.79\) & \(0.06\) & \(1.63\) & \(0.00097\) & \(0.17\%\)
& \(1.45\) & \(0.11\) & \(0.000030\) & \(0.84\) & \(0.1132\) & \(0.0176\%\) \\

Norm.\ error
& -- & \(0.84\%\) & \(0.08\%\) & \(0.64\%\) & \(<0.01\%\) & --
& -- & \(2.88\%\) & \(0.02\%\) & \(3.01\%\) & \(0.17\%\) & -- \\

\hline
\end{tabular*}
\end{table*}

\subsection{Determining Optimal Parameters via System Identification }

We first characterized the materials using stress-relaxation and creep-with-recovery experiments as shown in Fig.~\ref{Fig:SysID_combined}(c)--(f), following the system-identification procedure introduced in~\cite{Tolasa2026}. We measured stress relaxation by imposing a constant deformation of 5~mm and recording the resulting force decay for 3~s, and creep with recovery by applying a step force of 3~N for 3~s, abruptly reducing it to 0.5~N, and recording the corresponding deformation and recovery response. 

We used the stress relaxation measurements as the training data to identify the best fit of the IO-SLS, IO-GM, and FO-SLS models to the physical reference behavior. For the FO-SLS model, we set the short-memory length to \(N=101\), as this value provides good rendering performance with a feasible computational cost~\cite{Tolasa2026, Gemalmaz2026}.  

Table~\ref{tab:hil_sysid_combined} reports the system-identified model parameters and their corresponding fit quality for both materials. Fig.~\ref{Fig:SysID_combined} presents the best fit of the IO-SLS, IO-GM, and FO-SLS models to the stress relaxation of the physical materials.

In Table~\ref{tab:hil_sysid_combined}, $K_0$ is negative for the FO-SLS model, as this negative stiffness term compensates for the large positive stiffness introduced by the fractional-order element. The identified parameters produce a passive model, as established in~\cite{Gemalmaz2026}, indicating a physically realizable material model. 

Furthermore, we validated the system-identified rendering parameters through creep-with-recovery experiments. For the memory foam, the FO-SLS model achieved the lowest NRMSE of \(3.63\%\), compared with \(5.28\%\) for both the IO-SLS and IO-GM models. Similarly, for the hard silicone, the FO-SLS model yielded the lowest NRMSE of \(3.21\%\), whereas the IO-SLS and IO-GM models resulted in NRMSE values of \(5.73\%\) and \(5.74\%\), respectively.

\subsection{Determining Perceptually Optimal Parameters}

We used six population-level perceptually optimized parameter sets corresponding to the three rendering models and two reference materials. The two FO-SLS parameter sets were obtained from the HiL optimization experiments reported in the previous study~\cite{Tolasa2026}, whereas four additional experiments were conducted for the present study to determine the corresponding IO-SLS and IO-GM parameter sets. 

In each HiL optimization, participants evaluated the similarity of a rendered interaction to the physical reference using the ordinal categories of \emph{Close}, \emph{Similar}, and \emph{Different}. We used this ordinal feedback to iteratively update a Gaussian process~(GP) model of perceived realism and select subsequent rendering parameters. The experimental procedure is detailed in Section~\ref{sec:HiL_experiment}.

We kept the optimization framework and parameter settings the same as in~\cite{Tolasa2026} to leverage the already validated GP model hyperparameters and experimental conditions. In particular, the GP model used an RBF kernel, $k(\mathbf{x}_i,\mathbf{x}_j)=\exp\left(-\theta \lVert \mathbf{x}_i-\mathbf{x}_j\rVert^2/2\right)$, with \(\theta=30\). Ordinal feedback was modeled using thresholds \(t=\{-\infty,-0.5,0.5,\infty\}\), and new samples were selected using an upper confidence bound acquisition function with exploration weight \(\lambda=0.7\). Each model-specific optimization was conducted within a passivity-constrained and effective-stiffness-matched search space~\cite{Gemalmaz2026}. The search space was normalized to \([0,1]\) for GP modeling.

\begin{figure*}[t!]
    \centering
    \includegraphics[width=.95\textwidth]{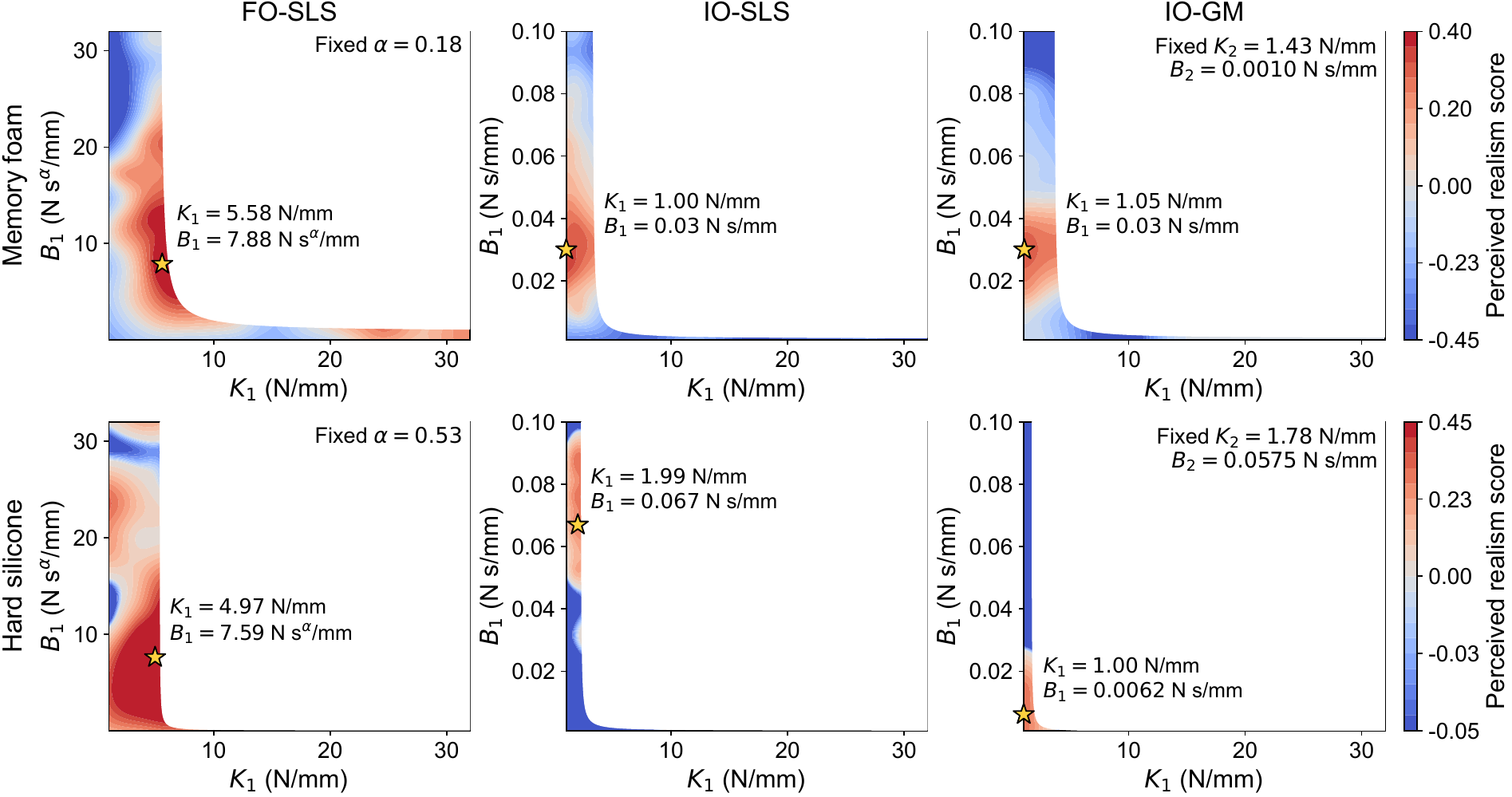}
    \vspace{-1.mm}
    \caption{Slices from aggregate posterior GP mappings of perceived realism. Each mapping shows a slice with the population-level optimized rendering parameters in the $K_1-B_1$ domain, with other parameters for the FO-SLS and IO-GM models fixed to the HiL-optimized values. Columns represent the rendering models; each row contains mappings from the same reference.} 
    \label{Fig:Aggregate}
\end{figure*}

After training the participant-specific GP models, we combined them into a single aggregated GP model using the Bayesian Committee Machine~\cite{Tresp2000} to obtain a population-level perceptual map for each viscoelastic model. We then selected the population-level perceptually optimized parameter set as the maximizer of the aggregate GP posterior mean, which represents perceived realism scores, as depicted in Fig.~\ref{Fig:Aggregate}. The aggregate optima represent generalized rendering parameters estimated to be perceived as most realistic by a group of users, rather than tailored to a single participant. 

Table~\ref{tab:hil_sysid_combined} presents the population-level perceptually optimal model parameters for both materials, along with the system-identified model parameters for ease of comparison.

\section{Human-Subject Experiments}

We first conducted HiL optimization sessions to determine population-level perceptually optimized parameters for the rendering models. We then used these parameters, along with the corresponding system-identified parameter sets, in the performance-comparison experiments to evaluate the perceived realism of the different model structures.

For both types of experiments, we recruited participants with no known sensory-motor impairments. We obtained written informed consent from all participants in accordance with the ethical guidelines approved by the Institutional Review Board of Sabanci University (Protocol No. FENS-2025-17).

\subsection{Apparatus}
\label{sec:apparatus}

\begin{figure}[t!]
    \centering
    \includegraphics[width=.75\linewidth]{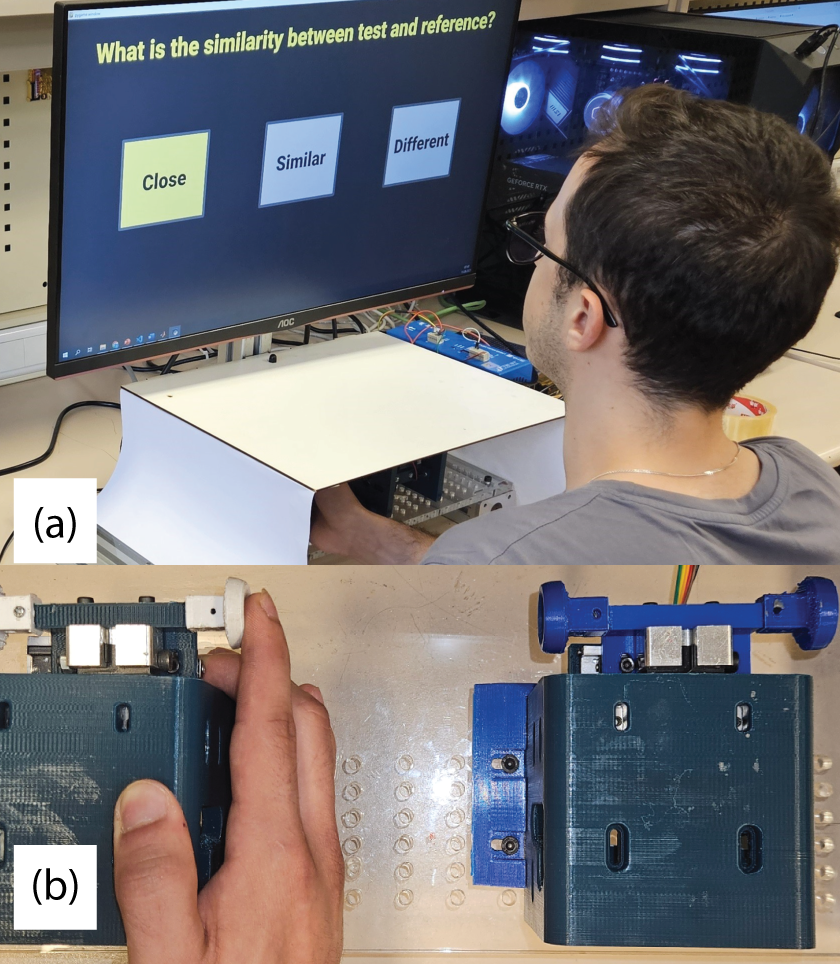}
    \vspace{-.5mm}
    \caption{(a)~The setup with a participant. (b)~Two identical haptic interfaces used for the human-subject experiments.}
    \label{fig:Haptic Device}
\end{figure}

The setup consisted of two identical haptic interfaces and a visual user interface, as in Fig.~\ref{fig:Haptic Device}. One device contained the physical viscoelastic material and served as the reference, while the other rendered a viscoelastic model with a chosen parameter set. Participants used their dominant hands, with the palm facing inward, and interacted using their index fingers.

Each haptic interface consists of a BEI Kimco LA28-15-002Z linear actuator and a U.S. Digital optical encoder with 2000 counts/inch resolution. The handle and supporting structure were 3D printed for comfortable grasping and repeatable interaction. We implemented open-loop impedance control in real-time at 1~kHz using the MATLAB~RTE.

\subsection{Participant Familiarization}

Before the experimental trials of both HiL optimization and performance-comparison sessions, participants completed an approximately five-minute warm-up in which they explored the physical reference and several viscoelastic renderings until they became familiar with the devices and the evaluation task.

\subsection{HiL Optimization Sessions}
\label{sec:HiL_experiment}

To determine haptic rendering parameters that maximize perceived realism relative to the physical material, we conducted four HiL optimization experiments using the IO-SLS and IO-GM models across the two reference materials.
\vspace{1.5mm}

\subsubsection{Participants}
For the HiL optimization experiments with both IO models, we recruited 24 participants for the memory-foam reference sessions and 8 participants for the hard-silicone reference sessions. All participants completed their corresponding sessions.

\vspace{1.5mm}
\subsubsection{Task and Experimental Design}

During the HiL optimization experiments, participants first interacted with the physical reference material, then with a test rendering selected by the HiL Bayesian optimization framework, for a total of~8~s. Participants evaluated the perceptual similarity of each rendering to the physical reference using the ordinal categories \emph{Close}, \emph{Similar}, and \emph{Different}, defined as follows: \vspace{1mm}

\begin{itemize}
\item \emph{Close}: The rendering feels the same as or highly similar to the physical reference.
\item \emph{Similar}: The rendering is perceptually distinguishable but moderately resembles the physical reference.
\item \emph{Different}: The rendering feels clearly different from the physical reference.
\end{itemize}\vspace{1mm}

We used these responses to update a GP model of perceived realism over the feasible parameter space. All HiL optimization sessions were conducted over 25~trials.

\subsection{Performance-Comparison Experiments}
\label{sec:comparison_experiment}
The performance-comparison experiments directly compared the perceived realism of the FO-SLS, IO-SLS, and IO-GM renderings using the HiL-optimized and system-identified parameter sets for both reference materials.

\vspace{1mm}
\subsubsection{Participants}

Two participant groups completed the experiments. 28 participants performed the comparison experiment for the memory foam, while 16 participants completed the experiment with the hard silicone reference.

\vspace{1mm}
\subsubsection{Task and Experimental Design}

Participants compared virtual renderings generated using the FO-SLS, IO-SLS, and IO-GM models with the corresponding physical reference. Comparisons were performed separately for the HiL-optimized and system-identified parameter sets, so that each trial compared model structures only under the same parameter-selection approach.

For each parameter-selection condition, the three models were compared in all possible pairings: FO-SLS vs IO-SLS, FO-SLS vs IO-GM, and IO-SLS vs IO-GM. Each pair was repeated four times, resulting in 12 trials per condition and 24 comparisons per participant. The order of the HiL-optimized and system-identified blocks was randomized. 

Participants were instructed to consider the overall haptic interaction, including stiffness, damping, and time-dependent viscoelastic behavior, while ignoring non-haptic cues. No correctness feedback was provided.

\begin{figure*}[t!]
\centerline{\includegraphics[width=1\textwidth]{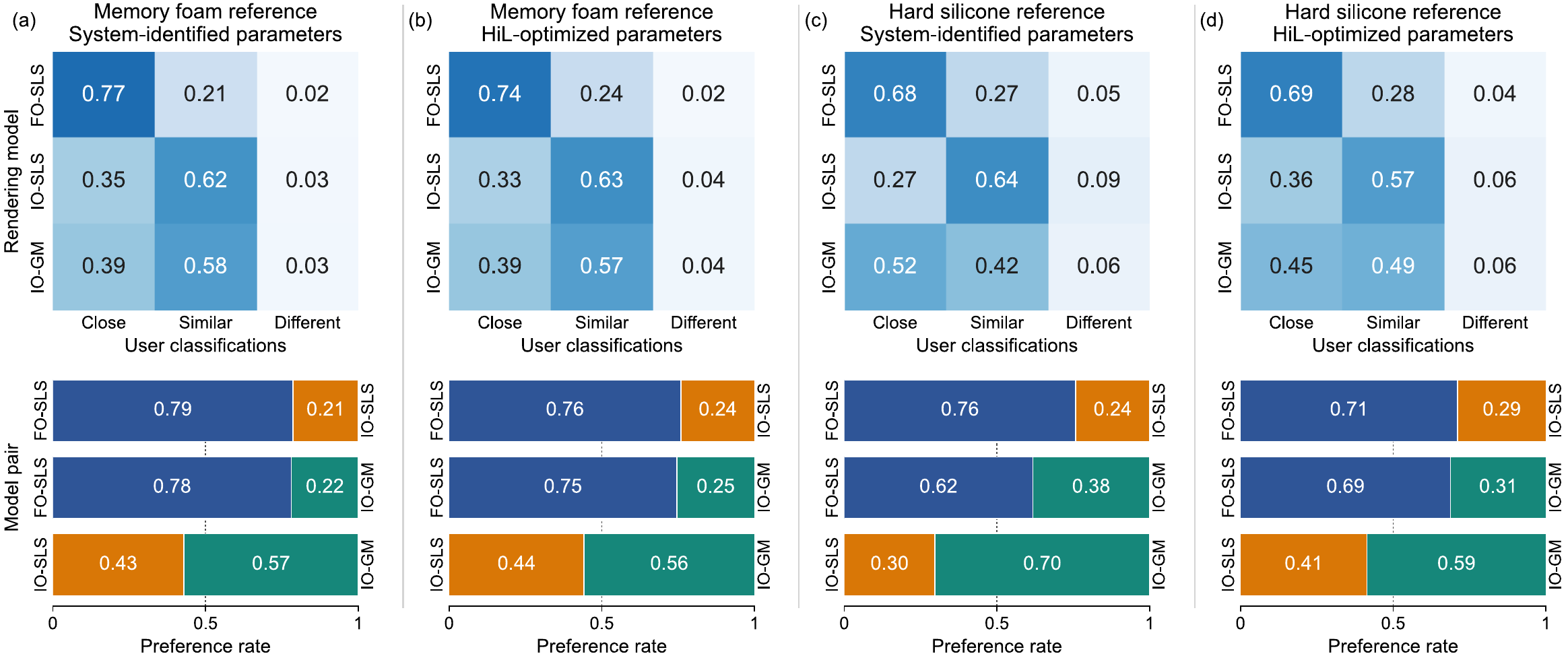}} 
\vspace{-1mm}
    \caption{Classification and pairwise preference results for the comparison between fractional- and integer-order viscoelastic renderings. The top row presents the classification-rate matrices, while the bottom row presents the corresponding pairwise preference percentages. The evaluations and comparisons for memory foam are shown in~(a) with system-identified rendering parameters~and in~(b) with HiL optimization parameters. Similarly, the results for hard silicone material are shown in~(c) for system-identified and in~(d) for HiL-optimized parameters.}
    \label{fig:fo_io_results}
\end{figure*}

\vspace{1mm}
\subsubsection{Experimental Procedure}

In each trial, participants first explored the physical reference using the reference device on the left and then interacted with two candidate virtual renderings using the test device on the right. The candidates were presented anonymously as \emph{A} and \emph{B} in randomized order. Participants were free to alternate between the two candidate renderings and were given a total of 12~s for exploration.

Following exploration, participants independently classified each rendering relative to the physical reference as \emph{Close}, \emph{Similar}, or \emph{Different} based on the same perceptual evaluation criteria used in the HiL optimization experiments.

After providing both classifications, participants selected the rendering that felt closer to the physical reference. All responses were recorded through the graphical user interface.


\section{Results and Discussion}

\begin{table*}[b]
\centering \small
\caption{Binary-logit analysis of the probability that each rendering is classified as \emph{Close} to the physical reference. Positive coefficients indicate a higher likelihood of a \emph{Close} classification.} 
\label{tab:close_logit}
\begin{tabular}{llcccc}
\textbf{Material} & \textbf{Rendering Model} &\textbf{Estimate} & \textbf{SE} &\textbf{$t$-value} & \textbf{$p$-value} \\
\hline \hline
\textbf{Memory foam}
& HiL FO       &  1.04 & 0.11 &  9.30 & $<.001$ \\
& SysID FO     &  1.20 & 0.14 &  8.65 & $<.001$ \\
& HiL IO-SLS   & -0.72 & 0.10 & -7.20 & 1.000 \\
& SysID IO-SLS & -0.61 & 0.09 & -6.56 & 1.000 \\
& HiL IO-GM    & -0.44 & 0.08 & -5.47 & 1.000 \\
& SysID IO-GM  & -0.44 & 0.09 & -4.93 & 1.000 \\
\hline \hline
\textbf{Hard silicone}
& HiL FO       &  0.79 & 0.09 &  8.95 & $<.001$ \\
& SysID FO     &  0.75 & 0.12 &  6.14 & $<.001$ \\
& HiL IO-SLS   & -0.56 & 0.11 & -4.94 & 1.000 \\
& SysID IO-SLS & -1.00 & 0.15 & -6.54 & 1.000 \\
& HiL IO-GM    & -0.20 & 0.17 & -1.24 & 1.000 \\
& SysID IO-GM  &  0.06 & 0.14 &  0.44 & 1.000 \\
\hline
\end{tabular} \normalsize 
\end{table*}

\subsection{Comparison of Perceived Realism of Viscoelastic Models}

We first evaluate the perceptual performance of the viscoelastic rendering models using classification and pairwise preference data collected during the performance comparison experiments. The results are summarized in Fig.~\ref{fig:fo_io_results} through classification matrices and pairwise preference rates.

For the memory foam, the FO-SLS model was classified as \emph{Close} to the physical reference in 77\% and 74\% of the evaluations using the system-identified and HiL-optimized parameter sets, respectively. In comparison, the corresponding \emph{Close} classification rates were 35\% and 33\% for the IO-SLS model and 39\% for the IO-GM model under both parameter-selection approaches. The pairwise preference results showed the same general trend. The FO-SLS model was preferred over the IO-SLS model in 79\% and 76\% of comparisons and over the IO-GM model in 78\% and 75\% of comparisons using the system-identified and HiL-optimized parameter sets, respectively.

Hard silicone reference also follows a similar trend. The FO-SLS model was classified as \emph{Close} in 68\% and 69\% of the evaluations using the system-identified and HiL-optimized parameter sets, respectively, compared to 27\% and 36\% for the IO-SLS and 52\% and 45\% for IO-GM models. In the pairwise evaluations, the FO-SLS model was preferred over the IO-SLS model in 76\% and 71\% of the comparisons and over the IO-GM model in 62\% and 69\% of the comparisons for the system identification and HiL optimization conditions, respectively. Overall, these results provide strong evidence that the FO-SLS model was evaluated more favorably than both IO model alternatives across the two reference materials and parameter-selection approaches.

To statistically evaluate these trends, we analyze the classification and pairwise preference responses using binary logistic regression models with cluster-robust standard errors to account for repeated responses within participants. For the classification analysis, the ordinal responses of \emph{Close}, \emph{Similar}, and \emph{Different} were converted into a binary outcome, with \emph{Close} coded as~1 and the remaining responses combined into a \emph{not Close} category, coded as~0. We fitted a binary logistic regression model separately for each rendering condition to test whether the probability of a \emph{Close} classification exceeded the reference probability of~$0.5$. Accordingly, a positive coefficient indicates that a model is more likely to be evaluated as \emph{Close}, whereas a negative coefficient indicates otherwise.

The results of the classification analysis are summarized in Table~\ref{tab:close_logit}. For the memory foam, both the HiL-optimized and system-identified FO-SLS renderings showed \emph{Close} classification probabilities above the reference level, with coefficients of $\beta=1.04$ ($t=9.30$, $p<0.001$) and $\beta=1.20$ ($t=8.65$, $p<0.001$), respectively. In contrast, the corresponding IO-SLS and IO-GM renderings yielded negative coefficients under both parameter-selection approaches. A similar trend was observed for the hard silicone, where the HiL-optimized and system-identified FO-SLS renderings yielded $\beta=0.79$ ($t=8.95$, $p<0.001$) and $\beta=0.75$ ($t=6.14$, $p<0.001$), respectively. None of the IO renderings showed evidence of exceeding the reference probability of a \emph{Close} classification.

The pairwise preference results support similar perceptual conclusions. As shown in Fig.~\ref{fig:fo_io_results}, the FO-SLS model was significantly preferred over both IO model alternatives for each reference material and under both parameter-selection strategies, with a clear dominance in preference rates, as also summarized in Table~\ref{tab:pairwise_logit}. For the memory foam, the FO-SLS rendering was preferred over IO-SLS and IO-GM renderings using both HiL-optimized parameters ($\beta=1.15$ and $\beta=1.07$, with $p<0.001$) and system-identified parameters ($\beta=1.30$ and $\beta=1.27$, with $p<0.001$). Similar results were observed for the hard silicone, where the FO-SLS rendering was preferred over IO-SLS and IO-GM renderings under both HiL optimization ($\beta=0.90$ and $\beta=0.79$, with $p<0.001$) and system identification ($\beta=1.14$ and $\beta=0.48$, with $p<0.001$).

The corresponding preference rates in Fig.~\ref{fig:fo_io_results} show that these differences are also substantial in magnitude. Depending on the material and parameter-selection strategy, the FO-SLS renderings were preferred over the IO-SLS renderings in $71$--$79\%$ of the comparisons and over the IO-GM renderings in $62$--$78\%$ of the comparisons.

Overall, both the classification and pairwise preference analyses indicate a consistent and clear perceptual advantage of the FO-SLS model over the two IO model alternatives. More importantly, this advantage was observed for both reference materials with diverse viscoelastic behavior and did not change when evaluating parameters determined through physical system identification or perceptual HiL optimization. Hence, the results strongly support that FO models not only provide a more representative mathematical model for capturing viscoelasticity but also achieve more realistic renderings than their IO counterparts.

\begin{table*}[t!]
\centering
\small
\caption{Pairwise preference analysis between the rendering conditions. 
Positive estimates indicate preference for the first rendering in each comparison, 
whereas negative estimates indicate preference for the second rendering. 
The preferred condition is reported only for statistically significant comparisons~($p<0.05$).} 
\label{tab:pairwise_logit}

\begin{tabular}{lllcccc l}

\textbf{Material} & \textbf{Analysis} & \textbf{Comparison} &
\textbf{Estimate} & \textbf{SE} & \textbf{$t$-value} &
\textbf{$p$-value} & \textbf{Preferred} \\
\midrule\midrule

\multirow{6}{*}{Memory foam}
& \multirow{3}{*}{HiL vs HiL}
& FO vs.\ IO-SLS    &  1.15 & 0.18 &  6.29 & $<0.001$ & FO \\
& & FO vs.\ IO-GM     &  1.07 & 0.14 &  7.59 & $<0.001$ & FO \\
& & IO-SLS vs.\ IO-GM & -0.23 & 0.13 & -1.85 & 0.065 & -- \\
\cmidrule(lr){2-8}

& \multirow{3}{*}{SysID vs SysID}
& FO vs.\ IO-SLS    &  1.30 & 0.15 &  8.83 & $<0.001$ & FO \\
& & FO vs.\ IO-GM     &  1.27 & 0.20 &  6.22 & $<0.001$ & FO \\
& & IO-SLS vs.\ IO-GM & -0.29 & 0.12 & -2.31 & 0.021 & IO-GM 
\\

\midrule \midrule

\multirow{6}{*}{Hard silicone}
& \multirow{3}{*}{HiL vs HiL}
& FO vs.\ IO-SLS    &  0.90 & 0.16 &  5.49 & $<0.001$ & FO \\
& & FO vs.\ IO-GM     &  0.79 & 0.19 &  4.12 & $<0.001$ & FO \\
& & IO-SLS vs.\ IO-GM & -0.35 & 0.21 & -1.66 & 0.096 & -- \\
\cmidrule(lr){2-8}

& \multirow{3}{*}{SysID vs SysID}
& FO vs.\ IO-SLS    &  1.14 & 0.20 &  5.67 & $<0.001$ & FO \\
& & FO vs.\ IO-GM     &  0.48 & 0.14 &  3.40 & $<0.001$ & FO \\
& & IO-SLS vs.\ IO-GM & -0.86 & 0.14 & -6.02 & $<0.001$ & IO-GM \\
\hline
\end{tabular} \normalsize
\end{table*}

\subsection{Comparison of Integer-Order Models}

Increasing the complexity of the IO representation nevertheless improved its ability to reproduce the measured viscoelastic response. As reported in Table~\ref{fig:visco_models}, the system-identified IO-GM model resulted in lower NRMSE than the IO-SLS model for both reference materials. For the memory foam, the NRMSE decreased from 1.42\% for the IO-SLS model to 0.17\% for the IO-GM model, while for the hard silicone it decreased from 0.062\% to 0.018\%. The improved physical fit obtained by adding a second Maxwell branch is also visible in Fig.~\ref{Fig:SysID_combined}.

This increase in modeling capability was reflected to some extent in the perceptual results. Under system identification, the IO-GM renderings were significantly preferred over the IO-SLS renderings for both the memory foam ($\beta=-0.29$, $p=0.021$) and hard silicone ($\beta=-0.86$, $p<0.001$). In contrast, no statistically significant preference between the two IO models was observed for the HiL-optimized parameter sets ($p=0.065$ and $p=0.096$ for the memory foam and hard silicone, respectively).

However, the additional complexity and improved physical fit of the IO-GM model were not sufficient to match the perceived realism of the FO-SLS model. The FO-SLS model was significantly preferred over the IO-GM model under both parameter-selection strategies and for both reference materials. This result is particularly informative for the system-identified memory-foam condition, where the IO-GM model achieved a marginally lower NRMSE  than the FO-SLS model (0.17\% versus 0.32\%), yet the FO-SLS model was preferred in 78\% of their pairwise comparisons. Therefore, improved agreement with the measured material response does not necessarily translate into improved perceived realism during haptic interaction.

\subsection{Comparison of Parameter-Selection Strategies}

Both system identification and HiL optimization provide complementary ways to determine rendering parameters. System identification selects parameters according to their ability to reproduce the measured physical response, whereas HiL optimization determines parameters directly from participants' evaluations of perceived realism. Despite these substantially different objectives, the resulting perceptual evaluations showed broadly similar trends.

This consistency is particularly apparent in the classification results as given in Fig.~\ref{fig:fo_io_results}. For the memory foam, the FO-SLS model received \emph{Close} classifications in $77\%$ and $74\%$ of the evaluations using the system-identified and HiL-optimized parameters, respectively, while the IO-SLS model received $35\%$ and $33\%$, and the IO-GM model received $39\%$ under both conditions. Similarly, for the hard silicone reference, the FO-SLS model had \emph{Close} rates of $68\%$ and $69\%$ for system-identified and HiL-optimized parameters. Furthermore, the pairwise comparisons also preserved the same overall model ordering across the two parameter-selection approaches.

These results are notable because the parameter sets obtained through system identification and HiL optimization are not necessarily identical, as indicated in Table~\ref{tab:hil_sysid_combined} and shown in prior studies~\cite{Catkin2023,Tolasa2026}. Nevertheless, the main perceptual trends were consistent across the two approaches. This suggests that, within the parameter sets considered in this study, the viscoelastic model structure influenced the overall perceptual ranking more than whether its parameters were selected through physical fitting or perceptual optimization.

\section{Conclusions}

We explored whether the enhanced physical fidelity offered by fractional-order viscoelastic modeling leads to perceptual advantages in haptic rendering. The human-subject validation experiments indicate that the FO-SLS model was consistently perceived as more realistic than both integer-order models used for comparison. In particular, when comparing the classification results, the FO-SLS model was rated as \emph{Close} to the physical reference significantly more frequently than the 50\% reference, whereas the IO-SLS and IO-GM models were rated as \emph{Close} significantly less frequently than this reference. The same trend appears in the pairwise preference results, where the FO-SLS model was significantly preferred over both the IO-SLS and IO-GM models for HiL-optimized and system-identified parameter sets. These findings show that the perceptual advantage of the FO-SLS model was robust across both perceptual and objective parameter-selection approaches for both viscoelastic reference materials.

The direct comparison between parameters from the HiL optimization and the system-identified parameters reveals that parameter selection matters less than the model structure. HiL optimization explicitly considers human perception, while system identification aims to best match the model to the measured response. Therefore, HiL optimization can enhance the perceived realism of more sensitive parameters. Nevertheless, the dominant trend in the validation experiments was that the FO-SLS model performed better than the tested integer-order variants. This indicates that the memory-dependent dynamics captured by the fractional-order formulation play a central role in realistically rendering viscoelastic behavior.

Overall, this study experimentally validates two complementary conclusions. First, FO-SLS viscoelastic models are not only concise and physically accurate descriptions of materials that exhibit creep and/or stress relaxation phenomena, but also perceptually advantageous models for haptic rendering. Second, human-in-the-loop optimization is a feasible and valid perceptual tuning approach, as it tunes model parameters based on user feedback rather than solely on the accuracy of the system identification process. 

Taken together, the results support the use of FO-SLS models for realistic viscoelastic virtual environments and suggest that fractional-order dynamics offer perceptual advantages that conventional integer-order models cannot fully recover, even when tuned through either system identification or human-in-the-loop optimization.

\section*{Acknowledgment}

This work has been partially supported by the TUBITAK Grant~23AG003. 

\bibliographystyle{IEEEtran}
\bibliography{Bibliography}

@article{tokatli_patoglu_2018,
title={Using Fractional Order Elements for Haptic Rendering}, 
journal={Springer Advanced Robotics Research}, 
author={Tokatli{}, Ozan and Patoglu, Volkan}, 
year={2018}, 
pages={373-–388}
}

@ARTICLE{Tolasa2024,
  author={Tolasa, Harun and Catkin, Bilal and Patoglu, Volkan},
  journal={IEEE Transactions on Haptics}, 
  title={Human-in-the-Loop Optimization of Perceived Realism of Multi-Modal Haptic Rendering Under Conflicting Sensory Cues}, 
  year={2025},
  volume={18},
  number={2},
  pages={295-311},
  doi={10.1109/TOH.2025.3535416}}

@article{Tolasa2026,
   title={Active Learning of Fractional-Order Viscoelastic Model Parameters for Realistic Haptic Rendering}, 
   author={Harun Tolasa and Gorkem Gemalmaz and Volkan Patoglu},
   year={2025},
   journal = {CoRR},
   url = {https://arxiv.org/abs/2512.00667},
}

@article{Catkin2023,
author = {Catkin, Bilal and Patoglu, Volkan},
journal = {IEEE Transactions on Haptics},
pages = {1--6},
title = {Preference-Based Human-in-the-Loop Optimization for Perceived Realism of Haptic Rendering},
year = {2023}
}

@inproceedings{Zhang2022,
  title={Assessing the Perceived Realism of Kinesthetic Haptic Renderings Under Parameter Variations},
  author={B. Zhang and M. Hagenow and B. Mutlu and M. Gleicher and M. Zinn},
  booktitle={IEEE Haptics Symposium},
  pages={1-6},
  year={2022},
}

@article{Swindells2009,
  title={Designing for feel: Contrasts between human and automated parametric capture of knob physics},
  author={C. Swindells and K. E. MacLean and K. S. Booth},
  journal={IEEE Transactions on Haptics},
  volume={2},
  number={4},
  pages={200--211},
  year={2009},
}

@article{prostate,
title = {Quantitative Characterization of Viscoelastic Properties of Human Prostate Correlated with Histology},
journal = {Ultrasound in Medicine \& Biology},
volume = {34},
number = {7},
pages = {1033-1042},
year = {2008},
issn = {0301-5629},
author = {Man Zhang and Priya Nigwekar and Benjamin Castaneda and Kenneth Hoyt and Jean V. Joseph and Anthony {di Sant'Agnese} and Edward M. Messing and John G. Strang and Deborah J. Rubens and Kevin J. Parker},
}

@article{breast_tissue,
  author       = {Carmichael, B. and Babahosseini, H. and Mahmoodi, S. N. and Agah, M.},
  title        = {The fractional viscoelastic response of human breast tissue cells},
  journal      = {Physical Biology},
  year         = {2015},
  volume       = {12},
  number       = {4},
  pages        = {046001},
  doi          = {10.1088/1478-3975/12/4/046001}
}

@article{kelvin-maxwell,
  title={Fractional generalizations of {Maxwell} and {Kelvin-Voigt} models for biopolymer characterization},
  author={J{\'o}{\'z}wiak, Bertrand and Orczykowska, Magdalena and Dziubi{\'n}ski, Marek},
  journal={PloS one},
  volume={10},
  number={11},
  pages={e0143090},
  year={2015},
  publisher={Public Library of Science San Francisco, CA USA}
}

@article{kelvin,
  title={{Kelvin-Voigt} versus fractional derivative model as constitutive relations for viscoelastic materials},
  author={Eldred, Lloyd B and Baker, William P and Palazotto, Anthony N},
  journal={AIAA journal},
  volume={33},
  number={3},
  pages={547--550},
  year={1995}
}

@article{frac-intg_model,
  title={A comparison between fractional-order and integer-order differential finite deformation viscoelastic models: Effects of filler content and loading rate on material parameters},
  author={Khajehsaeid, Hesam},
  journal={International Journal of Applied Mechanics},
  volume={10},
  number={09},
  pages={1850099},
  year={2018},
  publisher={World Scientific}
}

@article{frac_sls,
  title={Fractional calculus applied to model arterial viscoelasticity},
  author={Craiem, DO and Rojo, FJ and Atienza, JM and Guinea, GV and Armentano, Ricardo L},
  journal={Latin American Applied Research},
  volume={38},
  number={2},
  pages={141--145},
  year={2008}
}

@article{short_memory,
  title={Time-domain evaluation of fractional order controllers’ direct discretization methods},
  author={Ma, Chengbin and Hori, Yoichi},
  journal={IEEJ Transactions on Industry Applications},
  volume={124},
  number={8},
  pages={837--842},
  year={2004},
  publisher={The Institute of Electrical Engineers of Japan}
}

@article{gmm-sls,
  title={An equivalence between generalized {Maxwell} model and fractional {Zener} model},
  author={Xiao, Rui and Sun, Hongguang and Chen, Wen},
  journal={Mechanics of Materials},
  volume={100},
  pages={148--153},
  year={2016},
  publisher={Elsevier}
}

@INPROCEEDINGS{grunwald,
  author={Chen, YangQuan and Petras, Ivo and Xue, Dingyu},
  booktitle={American Control Conference}, 
  title={Fractional order control - {A} tutorial}, 
  year={2009},
  volume={},
  number={},
  pages={1397--1411},
}

@article{frac_better,
  title={Fractional calculus-a different approach to the analysis of viscoelastically damped structures},
  author={Bagley, Ronald L and Torvik, Peter J},
  journal={AIAA Journal},
  volume={21},
  number={5},
  pages={741--748},
  year={1983}
}

@article{cagatay_liver,
  title={Effect of solution and post-mortem time on mechanical and histological properties of liver during cold preservation},
  author={Ayyildiz, Mehmet and Aktas, Ranan Gulhan and Basdogan, Cagatay},
  journal={Biorheology},
  volume={51},
  number={1},
  pages={47--70},
  year={2014},
  publisher={SAGE Publications Sage UK: London, England}
}

@article{kelvin_orj,
  title={On the elasticity and viscosity of metals},
  author={Thomson, William},
  journal={Proceedings of the Royal Society of London},
  number={14},
  pages={289--297},
  year={1865},
  publisher={The Royal Society London}
}

@article{voigt_orj,
  title={Ueber die innere Reibung der festen K{\"o}rper, insbesondere der Krystalle.},
  author={Voigt, Woldemar},
  journal={Abhandlungen der Koeniglichen Gesellschaft der Wissenschaften in Goettingen},
  volume={36},
  pages={3--48},
  year={1890}
}

@article{maxwell_orj,
  title={On the dynamical theory of gases},
  author={Maxwell, James Clerk},
  journal={Philosophical Transactions of the Royal Society of London},
  number={157},
  pages={49--88},
  year={1867},
  publisher={The Royal Society London}
}

@article{zener_orj,
  title={Elasticity and Anelasticity of Metals.},
  author={Zener, Clarence M and Siegel, Sidney},
  journal={The Journal of Physical Chemistry},
  volume={53},
  number={9},
  pages={1468--1468},
  year={1949},
  publisher={ACS Publications}
}

@article{wiechert_orj,
  title={Gesetze der elastischen Nachwirkung f{\"u}r constante Temperatur},
  author={Wiechert, Emil},
  journal={Annalen der Physik},
  volume={286},
  number={11},
  pages={546--570},
  year={1893},
  publisher={Wiley Online Library}
}

@inproceedings{salisbury1995haptic,
  title={Haptic rendering: Programming touch interaction with virtual objects},
  author={Salisbury, Kenneth and Brock, David and Massie, Thomas and Swarup, Nick and Zilles, Craig},
  booktitle={Symposium on Interactive 3D graphics},
  pages={123--130},
  year={1995}
}

@article{breast_cancer,
  title={Characterizing viscoelastic properties of human melanoma tissue using Prony series},
  author={Park, Seungman and Chien, Anna L and Brown, Isabelle D and Chen, Jingchun},
  journal={Frontiers in Bioengineering and Biotechnology},
  volume={11},
  pages={1162880},
  year={2023},
  publisher={Frontiers Media SA}
}

@inproceedings{tokatli2015,
  title={Stability of haptic systems with fractional order controllers},
  author={Tokatli, Ozan and Patoglu, Volkan},
  booktitle={IEEE/RSJ International Conference on Intelligent Robots and Systems},
  pages={1172--1177},
  year={2015}
}

@article{time_dependant_tissue,
  title={Noninvasive assessment of the rheological behavior of human organs using multifrequency {MR} elastography: {A} study of brain and liver viscoelasticity},
  author={Klatt, Dieter and Hamhaber, Uwe and Asbach, Patrick and Braun, J{\"u}rgen and Sack, Ingolf},
  journal={Physics in Medicine \& Biology},
  volume={52},
  number={24},
  pages={7281},
  year={2007},
  publisher={IOP Publishing}
}

@article{IO_fails_memory,
  title={The fractional viscoelastic response of human breast tissue cells},
  author={Carmichael, B and Babahosseini, H and Mahmoodi, SN and Agah, M},
  journal={Physical Biology},
  volume={12},
  number={4},
  pages={046001},
  year={2015},
  publisher={IOP Publishing}
}

@article{FO_few_param,
  title={Fractional viscoelastic models for power-law materials},
  author={Bonfanti, Alessandra and Kaplan, Jonathan Louis and Charras, Guillaume and Kabla, Alexandre},
  journal={Soft Matter},
  volume={16},
  number={26},
  pages={6002--6020},
  year={2020},
  publisher={Royal Society of Chemistry}
}

@article{Mainardi2010,
      title={An Historical Perspective on Fractional Calculus in Linear Viscoelasticity}, 
      author={Francesco Mainardi},
      journal = {CoRR},
       year={2010},
      url={https://arxiv.org/abs/1007.2959}, 
}

@article{BagleyTorvik1983,
  author  = {R.L. Bagley and P.J. Torvik},
  title   = {A theoretical basis for the application of fractional calculus to viscoelasticity},
  journal = {Journal of Rheology},
  year    = {1983},
  volume  = {27},
  number  = {3},
  pages   = {201--210}
}

@INPROCEEDINGS{Rosenberg1993,
  author={Rosenberg, L.B. and Adelstein, B.D.},
  booktitle={IEEE Research Properties in Virtual Reality Symposium}, 
  title={Perceptual decomposition of virtual haptic surfaces}, 
  year={1993},
  pages={46-53},
  doi={10.1109/VRAIS.1993.378264}}

@ARTICLE{Caldiran2019,
  author={Caldiran, Ozan and Tan, Hong Z. and Basdogan, Cagatay},
  journal={IEEE Transactions on Haptics}, 
  title={Visuo-Haptic Discrimination of Viscoelastic Materials}, 
  year={2019},
  volume={12},
  number={4},
  pages={438-450},
  doi={10.1109/TOH.2019.2924212}}

@INPROCEEDINGS{Caldiran2018,
  author={Caldiran, Ozan and Tan, Hong Z. and Basdogan, Cagatay},
  booktitle={IEEE Haptics Symposium}, 
  title={An investigation of haptic perception of viscoelastic materials in the frequency domain}, 
  year={2018},
  pages={222--228}}

@article{Culbertson2017,
  author  = {Culbertson, Heather and Kuchenbecker, Katherine J.},
  title   = {Importance of Matching Physical Friction, Hardness, and Texture in Creating Realistic Haptic Virtual Surfaces},
  journal = {IEEE Transactions on Haptics},
  year    = {2017},
  volume  = {10},
  number  = {1},
  pages   = {63--74},
  doi     = {10.1109/TOH.2016.2598751},
}

@article{Turchet2013,
  author  = {Turchet, Luca and Burelli, Paolo and Serafin, Stefania},
  title   = {Haptic Feedback for Enhancing Realism of Walking Simulations},
  journal = {IEEE Transactions on Haptics},
  year    = {2013},
  volume  = {6},
  number  = {1},
  pages   = {35--45},
  doi     = {10.1109/TOH.2012.51},
}

@article{Friesen2024,
  author  = {Friesen, Rebecca Fenton and Vardar, Yasemin},
  title   = {Perceived Realism of Virtual Textures Rendered by a Vibrotactile Wearable Ring Display},
  journal = {IEEE Transactions on Haptics},
  year    = {2024},
  volume  = {17},
  number  = {2},
  pages   = {216--226},
  doi     = {10.1109/TOH.2023.3304899},
}

@article{Srinivasan1995,
  author  = {Srinivasan, Mandayam A. and LaMotte, Robert H.},
  title   = {Tactual Discrimination of Softness},
  journal = {Journal of Neurophysiology},
  year    = {1995},
  volume  = {73},
  number  = {1},
  pages   = {88--101},
  doi     = {10.1152/jn.1995.73.1.88},
}

@article{BergmannTiest2009,
  author  = {{Bergmann Tiest}, Wouter M. and Kappers, Astrid M. L.},
  title   = {Cues for Haptic Perception of Compliance},
  journal = {IEEE Transactions on Haptics},
  year    = {2009},
  volume  = {2},
  number  = {4},
  pages   = {189--199},
  doi     = {10.1109/TOH.2009.16},
}

@article{Jones2013,
  author  = {Jones, Lynette A. and Tan, Hong Z.},
  title   = {Application of Psychophysical Techniques to Haptic Research},
  journal = {IEEE Transactions on Haptics},
  year    = {2013},
  volume  = {6},
  number  = {3},
  pages   = {268--284},
  doi     = {10.1109/TOH.2012.74},
}

@article{Okamoto2013,
  author  = {Okamoto, Shogo and Nagano, Hikaru and Yamada, Yoji},
  title   = {Psychophysical Dimensions of Tactile Perception of Textures},
  journal = {IEEE Transactions on Haptics},
  year    = {2013},
  volume  = {6},
  number  = {1},
  pages   = {81--93},
  doi     = {10.1109/TOH.2012.32},
}

@article{Hover2010,
  author  = {H{\"o}ver, Raphael and Di Luca, Massimiliano and Harders, Matthias},
  title   = {User-Based Evaluation of Data-Driven Haptic Rendering},
  journal = {ACM Transactions on Applied Perception},
  year    = {2010},
  volume  = {8},
  number  = {1},
  pages   = {1--23},
  doi     = {10.1145/1857893.1857900},
}

@article{Yim2016,
  author  = {Yim, Sunghoon and Jeon, Seokhee and Choi, Seungmoon},
  title   = {Data-Driven Haptic Modeling and Rendering of Viscoelastic and Frictional Responses of Deformable Objects},
  journal = {IEEE Transactions on Haptics},
  year    = {2016},
  volume  = {9},
  number  = {4},
  pages   = {548--559},
  doi     = {10.1109/TOH.2016.2571690},
}

@article{Cha2022,
  author  = {Cha, Hojun and Bhardwaj, Amit and Choi, Seungmoon},
  title   = {Data-Driven Haptic Modeling and Rendering of Viscoelastic Behavior Using Fractional Derivatives},
  journal = {IEEE Access},
  year    = {2022},
  volume  = {10},
  pages   = {130894--130907},
  doi     = {10.1109/ACCESS.2022.3230065},
}

@article{Gemalmaz2026,
      title={Haptic Rendering of Fractional-Order Viscoelasticity: Passivity and Rendering Fidelity}, 
      author={Gemalmaz, Gorkem and Tolasa, Harun and Patoglu, Volkan},
      journal = {CoRR},
      year = {2026},
      url={https://arxiv.org/abs/2605.16389}, 
}

@ARTICLE{Tolasa_Pareto,
  author={Tolasa, Harun and Patoglu, Volkan},
  journal={IEEE Transactions on Haptics}, 
  title={Human-in-the-Loop Pareto Optimization: Trade-Off Characterization for Assist-as-Needed Training and Performance Evaluation}, 
  year={2026},
  volume={19},
  number={2},
  pages={339--352}
  }

@article{Tresp2000,
    author = {Tresp, Volker},
    title = "{A Bayesian Committee Machine}",
    journal = {Neural Computation},
    volume = {12},
    number = {11},
    pages = {2719-2741},
    year = {2000}
    }

\vspace{-.0\baselineskip}
\begin{IEEEbiography}[{\includegraphics[width=.8in,height=1.2in,clip,keepaspectratio]{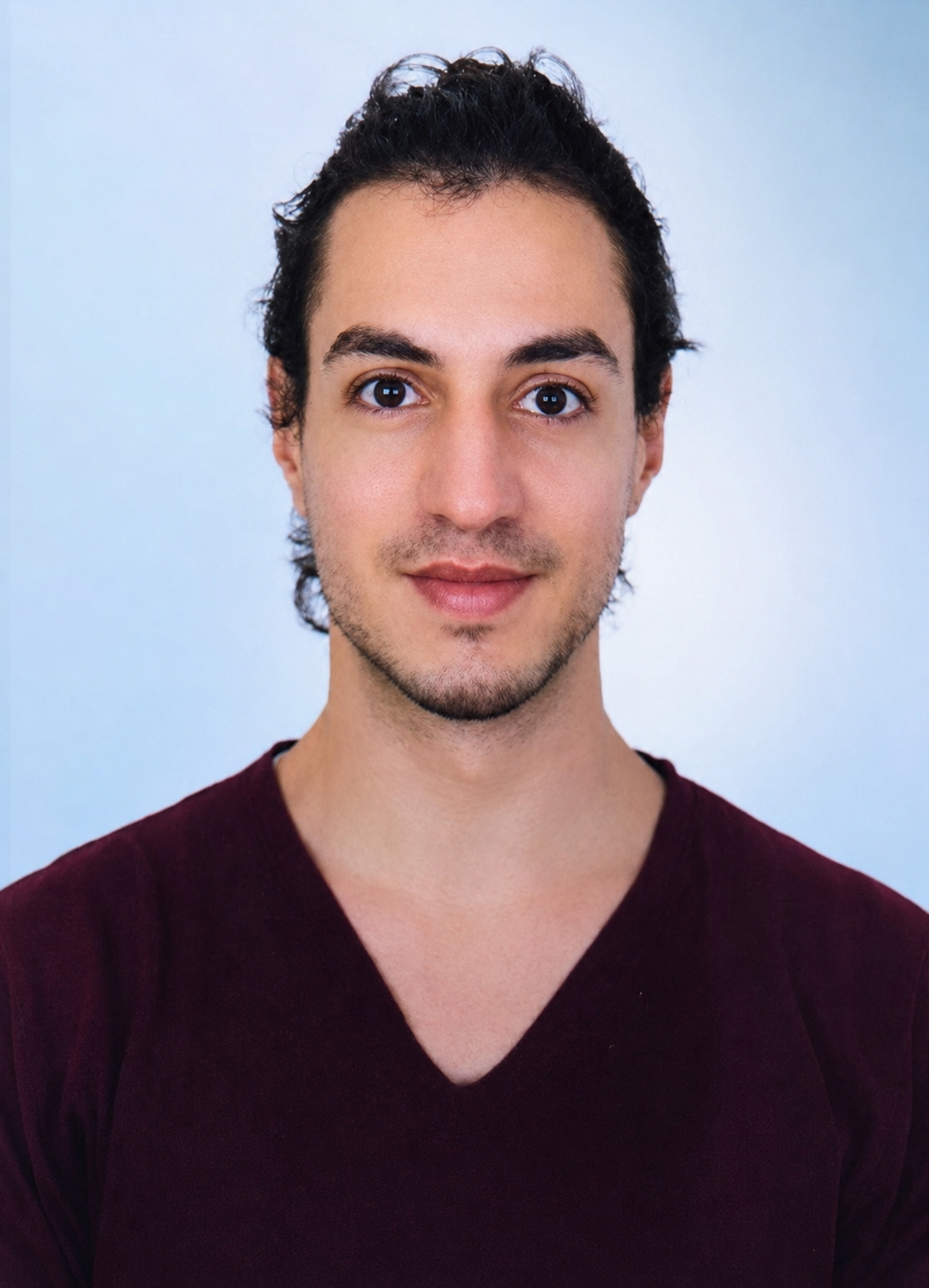}}]
{Gorkem Gemalmaz} received his B.Sc. in mechanical engineering from Middle East Technical University~(2022) and his M.Sc. in mechatronics engineering from Sabanci University~(2026). He is currently pursuing his Ph.D. at Sabanci University. His research interests include physical human-robot interaction and interaction control.
\end{IEEEbiography}

\vspace{-.0\baselineskip}
\begin{IEEEbiography}[{\includegraphics[width=.8in,height=1.2in,clip,keepaspectratio]{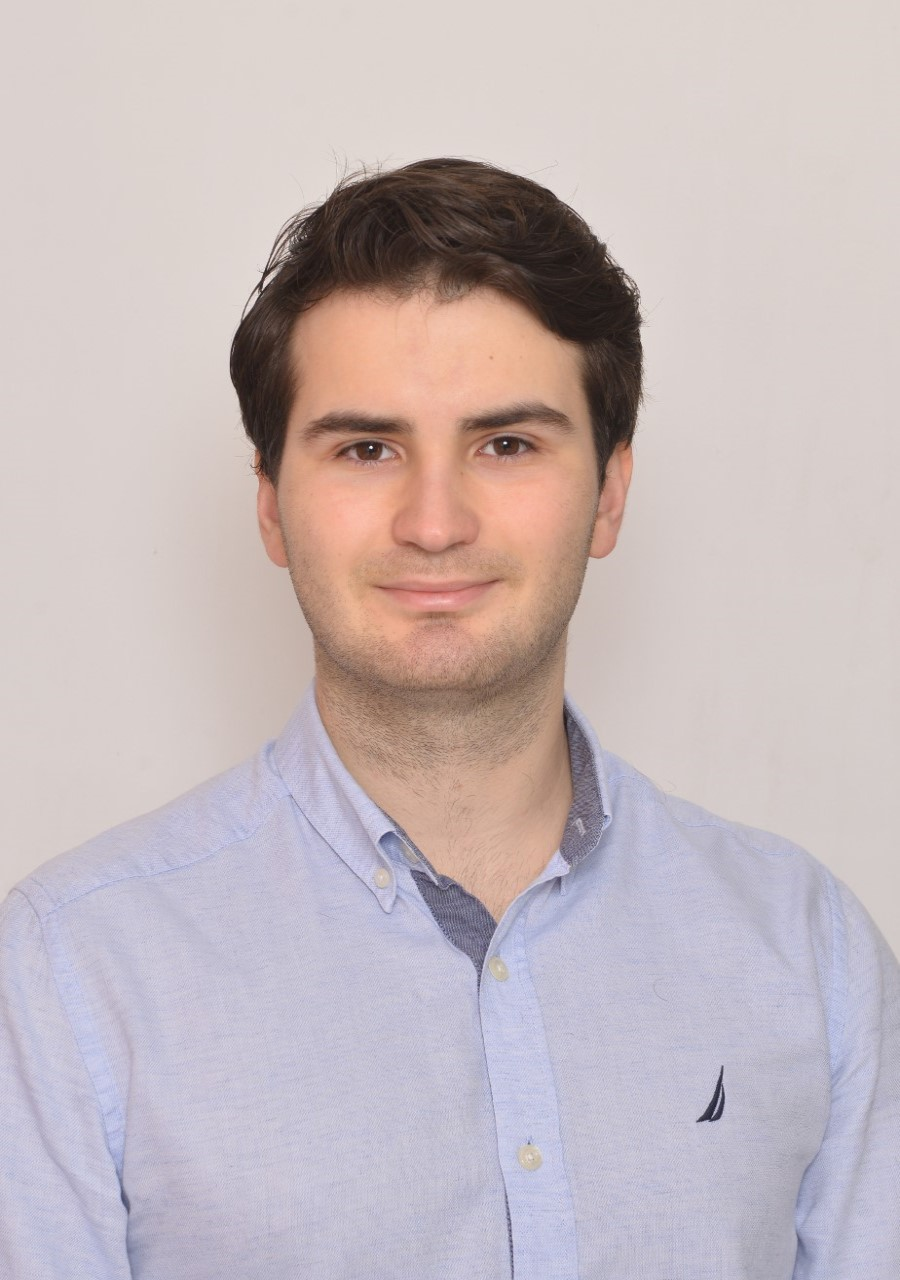}}]
{Harun Tolasa}  received his B.Sc. degree in mechanical engineering from Bilkent University (2021) and his M.Sc. in mechatronics engineering from Sabanci University~(2024). He is currently pursuing his Ph.D. at Sabanci University. His research interests include active learning, human-in-the-loop optimization, and haptic rendering.
\end{IEEEbiography}

\vspace{-.0\baselineskip}
\begin{IEEEbiography}[{\includegraphics[width=1in,height=1.2in,clip,keepaspectratio]{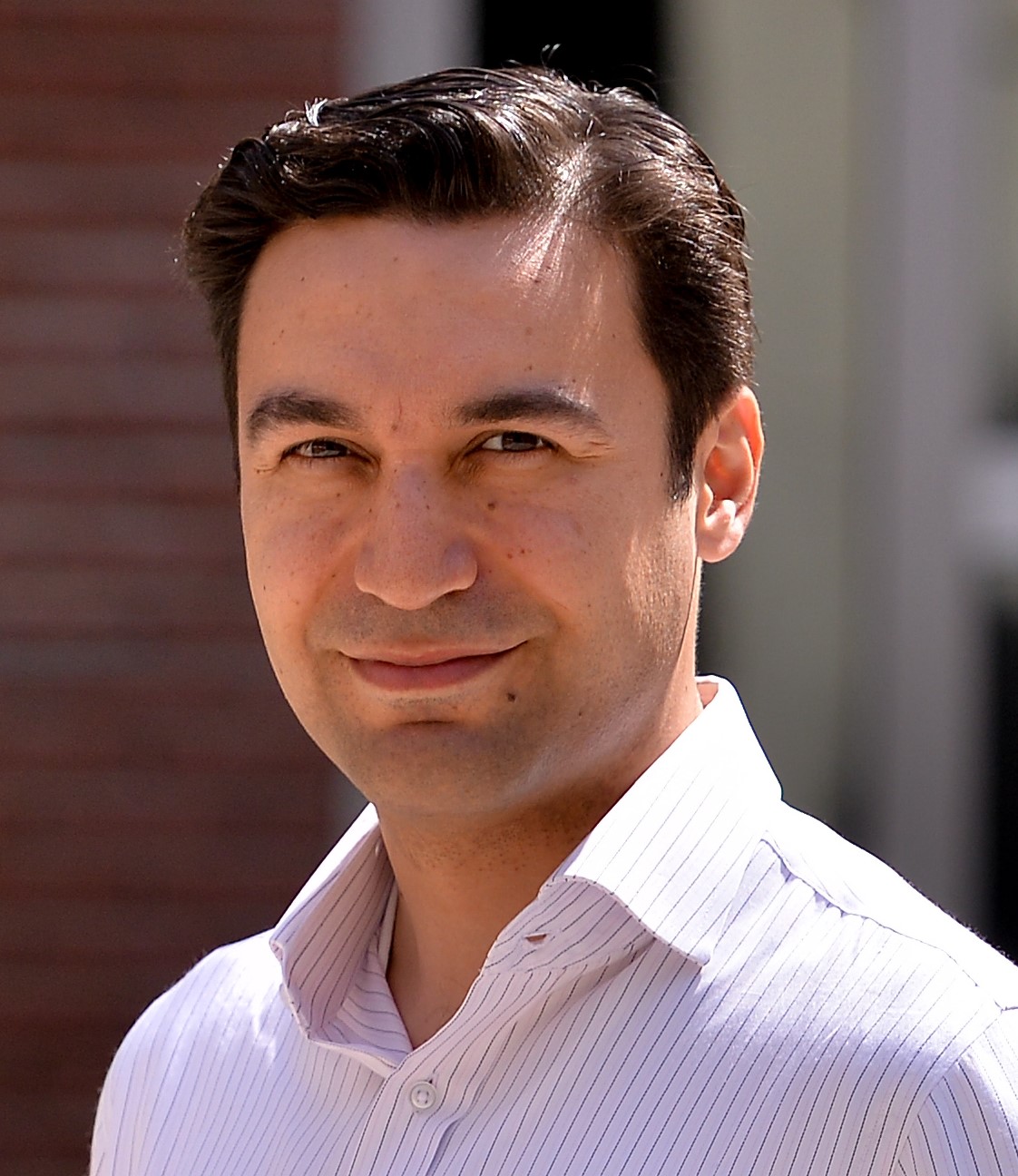}}]
{Volkan Patoglu} is a professor in mechatronics engineering at Sabanci University.
He received his Ph.D. in mechanical engineering from the University of Michigan, Ann Arbor~(2005) and worked as a postdoctoral researcher at Rice University~(2006). His research focuses on physical human-machine interaction, particularly the design and control of force-feedback robotic systems for rehabilitation applications. His research extends to cognitive robotics. He has served as associate editor for IEEE Transactions on Haptics~(2013--2017), IEEE Transactions on Neural Systems and Rehabilitation Engineering~(2018--2023), and IEEE Robotics and Automation Letters~(2019--2024).
\end{IEEEbiography}

\vfill

\end{document}